\documentclass[12pt, twocolumn]{article}
\usepackage[left=0.6in, right=0.6in, top=0.8in, bottom=0.8in]{geometry}
\usepackage{graphicx}
\usepackage{amsmath, amssymb}
\usepackage{hyperref}
\usepackage[numbers,sort&compress]{natbib}
\usepackage{titlesec} 
\usepackage{ragged2e}
\usepackage[T1]{fontenc}
\titleformat{\section}{\large\bfseries}{\thesection}{1em}{}
\titleformat{\subsection}{\normalsize\bfseries}{\thesubsection}{1em}{}
\newcounter{numquote}
\newenvironment{lquote}{
  \refstepcounter{numquote}
  \quote
}{
  \endquote
}

\title{Evaluating the Performance and Robustness of LLMs in Materials Science Q\&A and Property Predictions}
\author{
    Hongchen Wang$^{1}$, Kangming Li$^{2}$, Scott Ramsay$^{1}$,\\
    Yao Fehlis$^{3}$, Edward Kim$^{1,*}$, Jason Hattrick-Simpers$^{1,*}$
}
\date{March 2025}
\begin{document}
\justifying
\twocolumn[
    \maketitle
    \begin{center}
        {\small 
        $^{1}$ Department of Materials Science and Engineering, University of Toronto, Canada.\\ $^{2}$ Acceleration Consortium, University of Toronto, Canada.\\$^{3}$ Artificial, Inc., Austin, Texas, USA.\\$^{*}$ Corresponding authors: edwardsoo.kim@mail.utoronto.ca, jason.hattrick.simpers@utoronto.ca
        }
    \end{center}
    
    \begin{abstract}
    Large Language Models (LLMs) have the potential to revolutionize scientific research, yet their robustness and reliability in domain-specific applications remain insufficiently explored. In this study, we evaluate the performance and robustness of LLMs for materials science, focusing on domain-specific question answering and materials property prediction across diverse real-world and adversarial conditions. Three distinct datasets are used in this study: 1) a set of multiple-choice questions from undergraduate-level materials science courses, 2) a dataset including various steel compositions and yield strengths, and 3) a band gap dataset, containing textual descriptions of material crystal structures and band gap values. The performance of LLMs is assessed using various prompting strategies, including zero-shot chain-of-thought, expert prompting, and few-shot in-context learning. The robustness of these models is tested against various forms of ‘noise’, ranging from realistic disturbances to intentionally adversarial manipulations, to evaluate their resilience and reliability under real-world conditions. Additionally, the study showcases unique phenomena of LLMs during predictive tasks, such as mode collapse behavior when the proximity of prompt examples is altered and performance recovery from train/test mismatch. The findings aim to provide informed skepticism for the broad use of LLMs in materials science and to inspire advancements that enhance their robustness and reliability for practical applications.
    \end{abstract}
    \vspace{1em}
]

\section{Introduction}
Large Language Models (LLMs) represent a significant advancement in the field of artificial intelligence and have been rapidly adopted for application in various scientific disciplines \cite{Clusmann2023}. With their ability to process and generate natural language, LLMs are potent tools for tasks like information retrieval, question and answering (Q\&A), and property predictions \cite{lei2024materialsscienceeralarge, Jablonka2023, Gupta2022}. Similar to traditional ML models, LLMs can require extensive data processing, large volumes of data, and massive compute resources to train \cite{naveed2024comprehensiveoverviewlargelanguage}. Despite these limitations, pretrained LLMs can be adapted to new tasks with few-shot examples via in-context learning (ICL), making them both cost-effective and rapid to deploy \cite{wu2023openiclopensourceframeworkincontext, Zhao2021, vacareanu2024wordsnumberslargelanguage}. In the context of materials science, where data acquisition can often be costly and time-consuming, leveraging ICL enables LLMs to efficiently prototype and generate predictive insights even in low-data settings \cite{Schmidt2019, patwa2024enhancinglowresourcellmsclassification, li2024largelanguagemodelsincontext}. Recent work has demonstrated that LLMs are capable of domain-specific Q\&A \cite{Zaki2023, balhorn2023doeschatgptknownatural,mirza2024largelanguagemodelssuperhuman}, materials property predictions \cite{rubungo2023llmproppredictingphysicalelectronic, qian2023largelanguagemodelsempower, Jablonka2024}, and information extraction from complex datasets \cite{Gupta2022,chiang2024llamplargelanguagemodel}. These studies demonstrate LLMs’ potential to serve as flexible and powerful analytical tools. 

However, the robustness of LLMs is a critical factor in their practical deployment, yet it remains an underexplored area, particularly in domain-specific applications such as materials science. Previous studies have shown that LLMs struggle to maintain predictive accuracy when the input distribution shifts, exhibiting poor generalization to out-of-distribution (OOD) test data and vulnerability to adversarial attacks \cite{niven2019probingneuralnetworkcomprehension, utama2020debiasingnlumodelsunknown, du2021interpretingmitigatingshortcutlearning}. These challenges highlight the need for systematic robustness evaluations to ensure LLM reliability in real-world scenarios. A key aspect of the robustness of LLMs is their sensitivity to prompt changes either due to innocuous or adversarial reasons \cite{zhao2024improvingrobustnesslargelanguage, gu2023robustnesslearningtaskinstructions}. Variations in how a query or instruction is formulated may cause a response to factually change \cite{zhao2024improvingrobustnesslargelanguage}. As an example, 0.1 nm and 1 Angstrom are equivalent but switching them in a prompt could result in different LLM predictions for the same task. Alternatively, the response of the LLM can be deliberately altered through intentional misinformation or misleading inputs \cite{gu2023robustnesslearningtaskinstructions}. These attributes are not only theoretical concerns but are critical for the reliable usage of LLMs as they become integrated into the materials science research and development pipeline. Given that LLMs generate outputs with indifference to truth \cite{Hicks2024}, thoroughly probing LLM prompt sensitivity would allow us to critically evaluate model performance in practical situations; providing informed skepticism for the broad use of LLMs in materials science.  

In this work, we conducted a holistic robustness analysis of commercial and open-source LLMs for materials science. While our primary analyses focus on pre-reasoning models due to their consistent single-pass inference structure, we also include a representative reasoning model (\textit{DeepSeek-R1}\cite{deepseekai2025deepseekr1incentivizingreasoningcapability}) in our initial benchmarking to contextualize performance differences. Reasoning models, such as \textit{DeepSeek-R1} and \textit{OpenAI-o1}\cite{zhong2024evaluationopenaio1opportunities}, dynamically adjust inference strategies based on problem complexity, introducing variability that complicates direct comparisons with single-pass models. To maintain consistency and isolate inherent robustness characteristics, we restrict our subsequent robustness and sensitivity analyses solely to pre-reasoning LLMs. Three distinct datasets of domain-specific Q\&A and materials property prediction were selected. First, we benchmarked LLMs of different sizes and release periods using prompt engineering to establish baseline and optimal performance boundaries. We then investigated the impact of various textual perturbations, ranging from realistic to adversarial, on LLM performance in materials science Q\&A. We then used the matbench\_steels dataset to show that even without fine-tuning, pretrained LLMs demonstrated enhanced predictive ability through few-shot ICL when presented with similar examples to the prediction task. Conversely, when provided with dissimilar examples during few-shot ICL, mode collapse behavior was observed, where the model generated identical outputs despite varying inputs. Counterintuitively, for the fine-tuned LLM examined on band gap predictions, supposedly adversarial perturbations like shuffling or randomization enhanced the model’s predictive capability. Interestingly, this enables the fine-tuned model to function effectively even with significantly truncated prompts. This train/test mismatch behavior, absent in traditional ML models, highlights a potential direction for distilling LLM-based predictive models.

\section{Methods}

\begin{figure*}
 \centering
 \includegraphics[height=8cm]{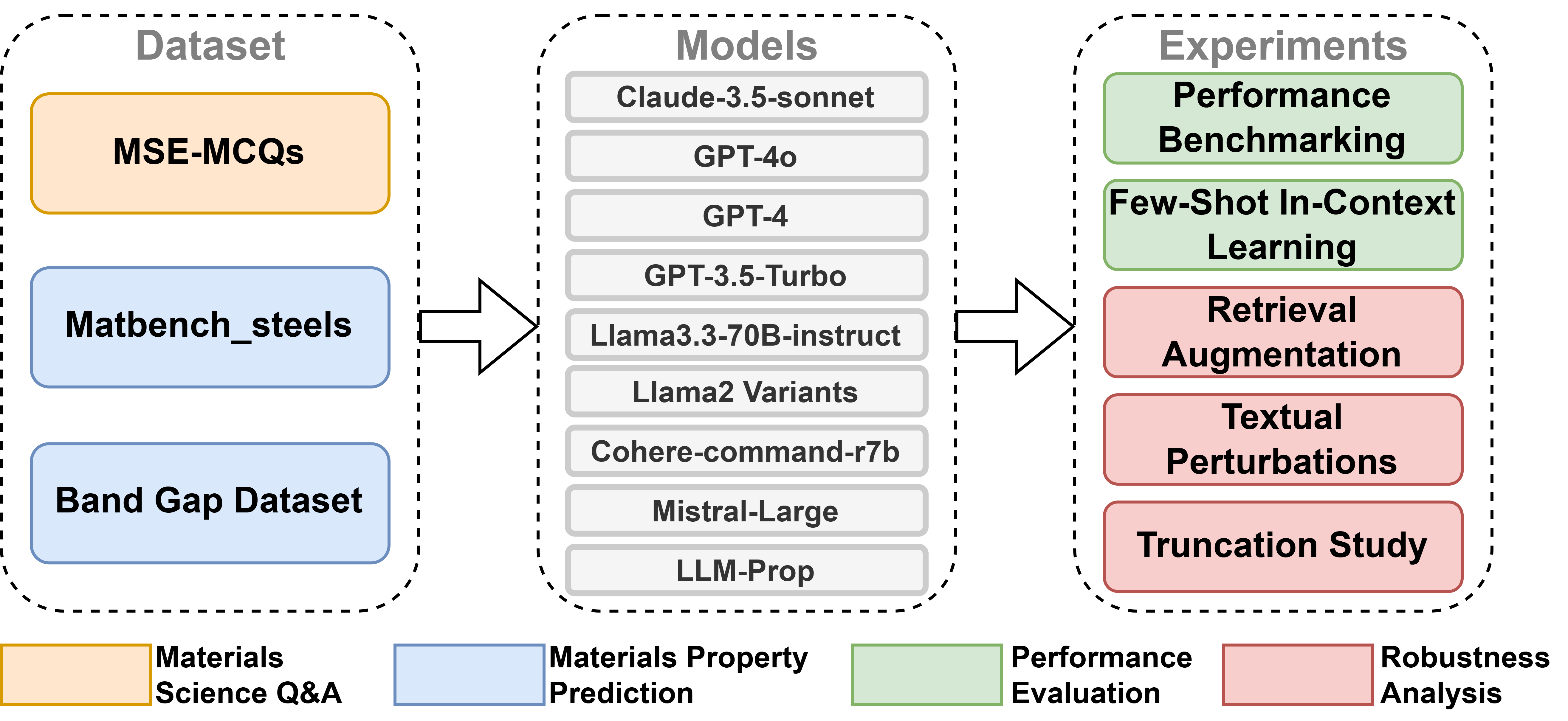}
 \caption{Schematic representation of the experiment design for performance evaluation and robustness analysis of various LLMs. Yellow highlights the testing conducted in Q\&A settings. Blue highlights the testing conducted in property prediction settings. Green represents the tests associated with performance evaluation. Red represents tests related to degradation and robustness analysis.}
 \label{Revision-figures/Revision-SchemeB.png}
\end{figure*}

The methodology is divided into four subsections that cover the performance evaluation and robustness analysis of LLMs in materials science Q\&A and property predictions. In each subsection, we will introduce the models, datasets used, prompting techniques, and evaluation criteria chosen for the specific study. All the evaluated models were set to their lowest temperature (typically 0) to minimize the non-determinism and maximize reproducibility. Figure \ref{Revision-figures/Revision-SchemeB.png} illustrates the experimental framework for evaluating LLMs in materials science Q\&A and materials property prediction. For performance evaluation and robustness analysis of materials science Q\&A, we compiled the MSE-MCQs dataset, consisting of 113 multiple-choice questions specifically designed for this study. These questions are original and were created by faculty at the University of Toronto for a first-year introductory materials science and engineering course. The questions were designed to test students’ understanding of materials science knowledge, including material mechanics, thermodynamics, crystal structures, materials properties, etc. The questions are manually categorized into easy (number of questions, n=39), medium (n=40), and hard levels (n=34), based on a set of heuristics, including conceptual complexity, the level of reasoning required, and the presence and difficulty of the calculations. For example, "easy" questions primarily test factual recall or direct application of basic concepts, such as identifying the crystal structure of a material. "Medium" questions involve moderate reasoning or straightforward calculations, such as determining the stress in a material under specific conditions. "Hard" questions require multi-step reasoning or more complex calculations, such as deriving material properties from combined thermodynamic and mechanical data. For the performance evaluation of property prediction, we use matbench\_steels, a subcategory of the Matbench test set originally proposed for benchmarking traditional machine learning (ML) models for materials property predictions \cite{Dunn2020}. The matbench\_steels dataset has 312 pairs of material compositions (as chemical formula) and yield strength (in MPa). For the robustness analysis of property prediction, we use a band gap dataset, which comprises 10,047 descriptions of material crystal structures generated via Robocrystallographer, along with their corresponding band gap values from the Materials Project database \cite{rubungo2023llmproppredictingphysicalelectronic}.

\subsection{Performance Evaluation of Materials Science Q\&A}

Using MSE-MCQs, we benchmarked a range of both commercial and open-source LLMs, including Anthropic's \textit{claude-3.5-sonnet-20240620}\cite{Anthropic2024}, OpenAI's \textit{gpt-4o-2024-11-20}\cite{openai2024gpt4ocard}, \textit{gpt-4-0613} \cite{openai2024gpt4technicalreport}, and \textit{gpt-3.5-turbo-0613} \cite{OpenAI2022}, alongside Meta AI’s Llama variants - \textit{llama3.3-70B-instruct}\cite{grattafiori2024llama3herdmodels}, \textit{llama2-70b-chat}, \textit{llama2-13b-chat}, and \textit{llama2-7b-chat} \cite{touvron2023llama2openfoundation}. The suffixes in the Llama model names denote the number of model parameters where, e.g., \textit{llama-70b} contains 70 billion parameters. A reasoning model, \textit{DeepSeek-R1}\cite{deepseekai2025deepseekr1incentivizingreasoningcapability} was also evaluated. This study primarily utilized fixed-version and open-source models to enhance the reproducibility of the findings.

The prompting strategies used in this evaluation are zero-shot chain-of-thought (CoT) and expert prompting. Zero-shot CoT prompting uses an instructional prompt to enable the model to "think aloud" and step through a reasoning process that leads to an answer, potentially boosting the accuracy in problem-solving tasks \cite{wei2023chainofthoughtpromptingelicitsreasoning}. Expert prompting, i.e., instructing an LLM to act as a domain expert, has been shown to influence the model's responses to be more aligned with expert knowledge and reasoning \cite{Xu2023}. The specific implementations of each for their respective tasks are defined below.

In the Q\&A evaluation, the expert prompt includes instructions to define the domain of study, introduces the settings of the questions, and emphasizes step-by-step reasoning and calculations. The goal is to improve the LLMs’ ability to retrieve domain-specific knowledge, follow the instructions, and correctly perform reasoning and calculations. The expert prompt is shown below:

\begin{lquote} 
\footnotesize
\textbf{Expert Prompt: }\textit{You are a renowned materials science engineering professor with extensive knowledge in the field. Your students have presented you with a challenging multiple-choice question related to materials science engineering. The question requires a detailed understanding and application of materials science principles. Please read the question carefully and provide a step-by-step explanation of your reasoning process, calculations, and analysis. Remember, the question has only one correct answer, which could be option (a), (b), (c), (d), etc. After carefully analyzing and calculating, please present the final answer at the end of your explanation. Your goal is to elucidate the concepts and problem-solving techniques in materials science engineering for your students.} 
\label{expert-prompt}
\end{lquote}

The questions were fed to LLMs in two settings: 1) as is, to benchmark the LLM’s level of understanding of materials science domain-specific knowledge and problem-solving; 2) to evaluate the aforementioned prompting strategies to enhance the LLM performance. Given the lengthy reasoning in the answers and the potential for errors in manual verification, we used the \textit{gpt-4-0613} API in a separate client to extract and assess responses automatically. This system compared answers to the provided correct choices, generating a simple binary score (1 for correct, 0 for incorrect) without evaluating the reasoning steps. Finally, the average accuracy of each category was calculated and reported. When selectively compared to manual checks (>200 answers), the method was found to be reliable, consistently identifying correct answers with over 95\% accuracy. The prompt is shown below: 

\begin{quote}
\footnotesize
\textbf{System Prompt: }\textit{You are to read the following text, which is the answer to a multiple choices question. The text should state the final answer (option (a), (b), (c), or (d)). You are to compare the stated answer with the correct answer: \textbf{ANSWER}. If the stated answer is correct, please type 1, otherwise type 0. If the final answer is not one of the options or reports multiple options, it is considered wrong (you should type 0). Do not type anything else other than 1 or 0.} 
\end{quote}
 
\subsection{Performance Evaluation of Materials Property Prediction}

For this study, we used the matbench\_steels dataset to evaluate the predictive capabilities of pretrained LLMs, using their ICL capability with few-shot examples. We benchmarked \textit{claude-3.5-sonnet-20240620}, \textit{gpt-4o-2024-11-20}, \textit{gpt-3.5-turbo-1106} \textit{llama3.3-70B-instruct}, \textit{mistral-large-2411}\cite{Mistral2024}, and Cohere's \textit{command-r7b-12-2024}\cite{Cohere2024} on their abilities to predict the yield strengths given the steel compositions. For performance comparison, a \textit{KNN} and an \textit{RFR} model were also implemented. 
Few-shot learning involves providing the LLM with a few examples of the task at hand, enabling it to learn the pattern and apply it to unseen questions or problems \cite{gao2021makingpretrainedlanguagemodels}. To use LLMs as predictive models, we fed the “training data” as few-shot examples to the prompt windows of the LLMs. An example of the “training data” prompt is shown below. 

\begin{quote}
\footnotesize
\textbf{System Prompt: }\textit{Given some example alloy compositions and their yield strengths, predict the yield strength for one additional alloy composition.}
\\\textbf{User Prompt: }\textit{\#\#\#Examples\#\#\# 
\\composition: Fe0.682 C0.00925 Mn0.000101 Si0.0101 Cr0.134 Ni0.00899 Mo0.0115 V0.000109 Nb0.000479 Co0.143 Al0.000618, yield strength: <1314.2>
\\composition: Fe0.792 C0.000470 Mn0.000411 Si0.00201 Cr0.0862 Ni0.0980 Mo0.0181 V0.000111 Nb0.0000607 Co0.0000957 Al0.00167 Ti0.000589, yield strength: <1061.7>
\\Now predict the yield strength for the following composition. 
\\composition: Fe0.768 C0.000931 Mn0.00244 Si0.00199 Cr0.110 Ni0.0981 Mo0.0113 V0.000110 Nb0.0000602 Co0.0000948 Al0.00497 Ti0.00269
\\Write only the yield strength in a single numerical value that is enclosed by <>} 
\end{quote}

Starting with an instruction, compositions were restructured by separating each element with a space and then paired to their corresponding yield strengths. We varied the number of training data points from 5 to 25 to observe how LLMs’ prediction accuracy scales with data size. Beyond 25 points, some models suffered from limited prompt windows. To assess the impact of data selection variability, five random seeds were chosen to vary the selected data.  To compare the predictive capability of LLMs and traditional ML models, we also utilized an \textit{RFR} model trained using the same data points. The \textit{RFR} model, unlike the LLMs that operated on raw compositional data, was trained with MAGPIE features \cite{Ward2016} extracted from the compositions, aiming to compare the predictive capability of LLMs and traditional ML models. The selected features are presented in the supplementary information. 
To evaluate the impact of the proximity of the few-shot examples on the predictive performance, a retrieval-augmented method was used. Specifically, the compositions were encoded by their elements and contents and then transformed the compositions into principal components (PCs). Given each test composition, we formulated its “training” set by its Euclidean distances (L2 norm) to the other points in the PC space. Three settings were chosen based on the distances: 1) random; 2) nearest neighbors; 3) farthest neighbors. The prediction accuracy, mean absolute error (MAE), in each setting was computed and analyzed.

\subsection{Robustness Analysis of Materials Science Q\&A}

To evaluate the robustness of LLMs for materials science Q\&A, we continued using the MSE-MCQs dataset and tested \textit{gpt-4o-2024-11-20} and \textit{gpt-3.5-turbo-0613} to observe how textual perturbations impacted performance. In this study, no prompting strategy was implemented, and the questions were directly used as the user prompts. 
We identified different types of “noise” that can be introduced to the MCQs to evaluate the robustness of LLMs. As shown in Table \ref{tab:degradation_types}, the textual inputs were modified systematically in five different ways, i.e., 1) unit mixing, 2) sentence reordering, 3) synonym replacement, 4) distractive info, and 5) superfluous info.

\begin{table*}[t]  
\footnotesize
\centering
\renewcommand{\arraystretch}{1.2} 
\caption{Types and descriptions of textual degradation applied to LLMs}
\label{tab:degradation_types}
\resizebox{\textwidth}{!}{  
\begin{tabular}{p{0.2\textwidth} p{0.4\textwidth} p{0.4\textwidth}} 
    \hline
    \textbf{Degradation Type} & \textbf{Description} & \textbf{Goal} \\
    \hline
    Unit Mixing & Mixing and converting the units & To test LLMs' interpretation of different unit systems and calculation abilities \\
    Sentence Reordering & Reordering the sentences in the questions & To assess LLMs’ capability to maintain comprehension on varied sentence constructions and logical flow \\
    Synonym Replacement & Replacing technical nomenclature with their synonyms & To evaluate the semantic understanding and stability of LLMs \\
    Distractive Info & Adding non-materials-science-related distractive information to the questions & To test LLMs’ ability to filter out irrelevant data \\
    Superfluous Info & Adding materials-science-related superfluous information containing numerical values to the questions & To challenge LLMs' ability to identify relevant data without being misled by additional numeric details \\
    \hline
\end{tabular}
}
\end{table*}

These modifications are expected to vary in their impact on the LLMs' performance, with some potentially degrading it due to their adversarial nature (such as reordering sentences and adding superfluous materials-science information) and others more realistically simulating conditions encountered in real-life scenarios. Considering the inherent variability due to the non-deterministic nature of LLMs, the test was repeated three times for the original, synonym replacement, and distractive info (same input texts). The unit mixing, sentence reordering, and superfluous info were randomized three times to introduce variability in the data for the evaluation. Finally, the accuracy of each category was calculated and reported.

\subsection{Robustness Analysis of Materials Property Prediction}

In materials property prediction, we selected the \textit{LLM-Prop} model along with its associated band gap dataset. \textit{LLM-Prop} is a fine-tuned T5 model, topped with a linear layer, designed to predict materials properties from crystal structure descriptions generated using Robocrystallographer \cite{rubungo2023llmproppredictingphysicalelectronic, Ganose2019, Raffel2020}.

The material descriptions underwent systematic modifications mirroring those applied in the Q\&A evaluations, except for unit mixing and synonym replacement. Note that, because of the highly templated nature of crystal structure descriptions, superfluous information in this context is better characterized as misleading information rather than simply extraneous text. During data preprocessing for \textit{LLM-Prop}, all numerical values and units, such as bonding distances and angles, are replaced with a [NUM] token, to emphasize the model's focus on text-based understanding \cite{rubungo2023llmproppredictingphysicalelectronic}. Unit mixing might disrupt the preprocessing algorithm, and thus was excluded from the analysis. Synonym replacement was excluded because the original terminology was already highly specific and lacked equivalent synonyms. Furthermore, we conducted a truncation study of textual degradation to examine the model's resilience against structural and length variations in the input data, as well as to explore which aspects of the descriptions the model relies on for predictions. We manipulated the order and fraction of sentences included, testing configurations including 1) original order, which prioritizes the initial information in a description, 2) reverse order, which prioritizes the sentences from the end of a description, 3) random order, shuffling the information, and 4) sides-to-middle, which deprioritized central information. The impact of these textual degradations was quantitatively assessed by measuring the resultant prediction error in MAE.

\section{Results and Discussion}

\subsection{Performance Evaluation of Materials Science Q\&A}

\begin{figure*}[h]
\centering
    \includegraphics[height=9cm]{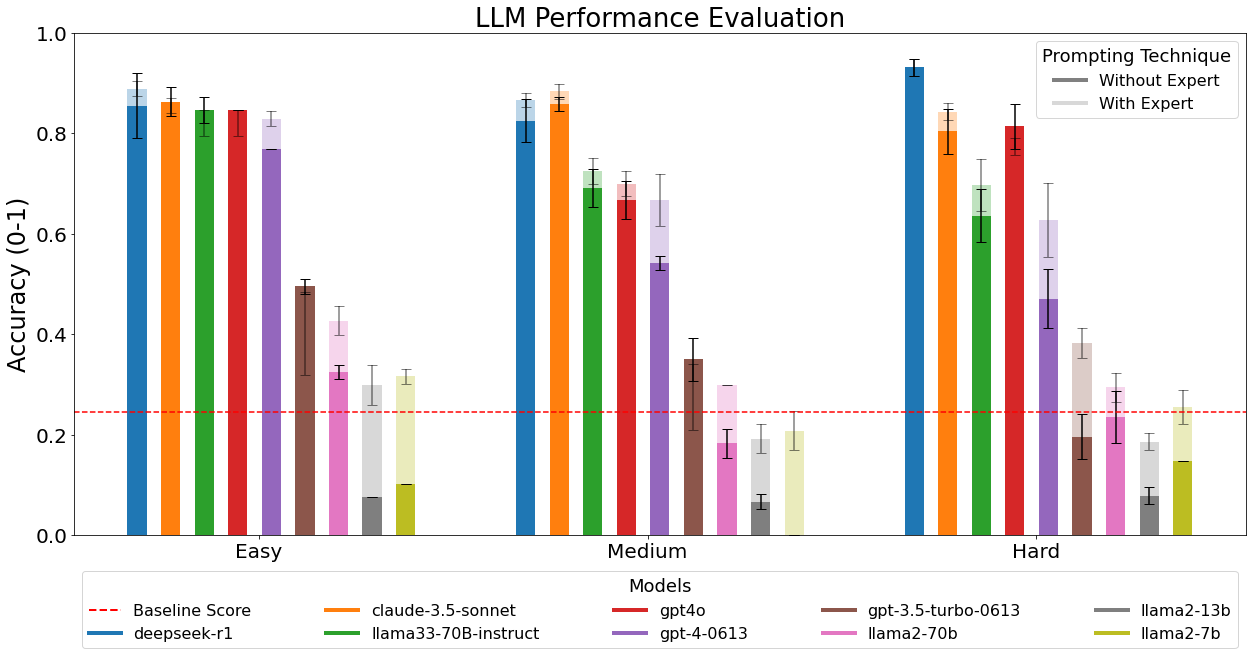}
    \caption{LLM performance evaluation and prompt engineering enhancement in materials science Q\&A using MSE-MCQs dataset. On each bar, the lighter color represents the performance of those models when introduced with the expert prompt. Error bars represent the standard deviation due to LLMs’ non-determinism.}
    \label{Revision-figures/Revision-llm_performD.png}
\end{figure*}

The results of the performance evaluation of LLMs on the MSE-MCQs dataset are shown in Figure \ref{Revision-figures/Revision-llm_performD.png}. For each model and setting, three trials were conducted, and the error bars represent the inherent non-determinism in LLMs even at the lowest temperature settings (typically 0). This non-determinism may affect the reliability and reproducibility of the models \cite{yu2023benchmarkinglargelanguagemodel} and is likely a result of stochastic sampling during text generation \cite{mitchell2023detectgptzeroshotmachinegeneratedtext, krishna2022rankgenimprovingtextgeneration, ouyang2023llmlikeboxchocolates}. We evaluated several advanced models released after late 2024 (i.e., \textit{DeepSeek-R1} (reasoning model), \textit{claude-3.5-sonnet-20240620}, \textit{llama3.3-70B-instruct}, and \textit{gpt-4o-2024-11-20}), along with some older models (i.e., \textit{gpt-4-0613}, \textit{gpt-3.5-turbo-0613}, and the \textit{llama2} variants). Overall, the newer and larger models significantly outperform older models, highlighting substantial recent advancements in LLM capabilities. 

Among the evaluated pre-reasoning models, \textit{claude-3.5-sonnet-20240620} achieved the highest accuracy across all difficulty levels. Notably, the reasoning model \textit{DeepSeek-R1} demonstrated competitive performance, closely matching \textit{claude-3.5-sonnet-20240620} on easy and medium questions, and surpassing all models on hard questions. The hard questions predominantly involve complex, multi-step reasoning or advanced mathematical calculations, tasks that typically present substantial challenges for single-pass pre-reasoning models. This superior performance by \textit{DeepSeek-R1}'s clearly highlights its inherent strength in tasks demanding deeper analytical and mathematical reasoning compared. 

The older \textit{llama2} models performed at or slightly below the baseline, equivalent to random guessing, while the newer \textit{llama3.3-70B-instruct} achieved comparable accuracy to \textit{gpt-4o-2024-11-20} on easy and medium questions.

Upon implementing the expert prompt (See \hyperref[expert-prompt]{Expert Prompt}), we observe consistent performance improvement across almost all models and question types. The improvement is more significant on the older models, suggesting that the expert prompt can enhance reasoning abilities with weaker baseline capabilities. However, the expert prompt provides minimal benefit for the newer pre-reasoning models on the easy questions, likely because the extensive reasoning process induced by the expert prompt contributes little to the performance on simple conceptual questions that rely primarily on factual recall. Interestingly, \textit{DeepSeek-R1} shows no performance improvement on hard questions upon implementing the expert prompt, suggesting that the reasoning capabilities of \textit{DeepSeek-R1} are already effectively saturated by its built-in iterative reasoning mechanism, such that additional explicit prompting does not further augment its performance.

We further investigated why the smaller \textit{llama2} models (\textit{llama-13b-chat} and \textit{llama-7b-chat}) scored lower than the baseline without the expert prompt. Despite being chat models, they sometimes failed to understand the intent when instructions were not provided. Instead of answering, they often attempted to “complete” the questions. Once the expert prompt was implemented, these smaller models could follow the instructions and attempt to solve the questions, in which case the performance improved to around and sometimes above the baseline scores. However, their overall performance remained weak due to their limited skill levels. 

Overall, the observed performance trends align with expectations: more recent and larger models consistently demonstrate enhanced capabilities in domain-specific Q\&A tasks compared to their predecessors. Additionally, prompt engineering demonstrated effectiveness as a strategy for enhancing model performance when handling more complex questions, especially for older or smaller models with limited baseline capabilities. On the other hand, the reasoning model, \textit{DeepSeek-R1}, exhibits inherently superior performance in complex analytical and mathematical reasoning tasks, achieving high accuracy even without specialized prompting.

\subsection{Performance Evaluation of Materials Property Prediction}

\begin{figure*}
\centering
\includegraphics[height=9.2cm]{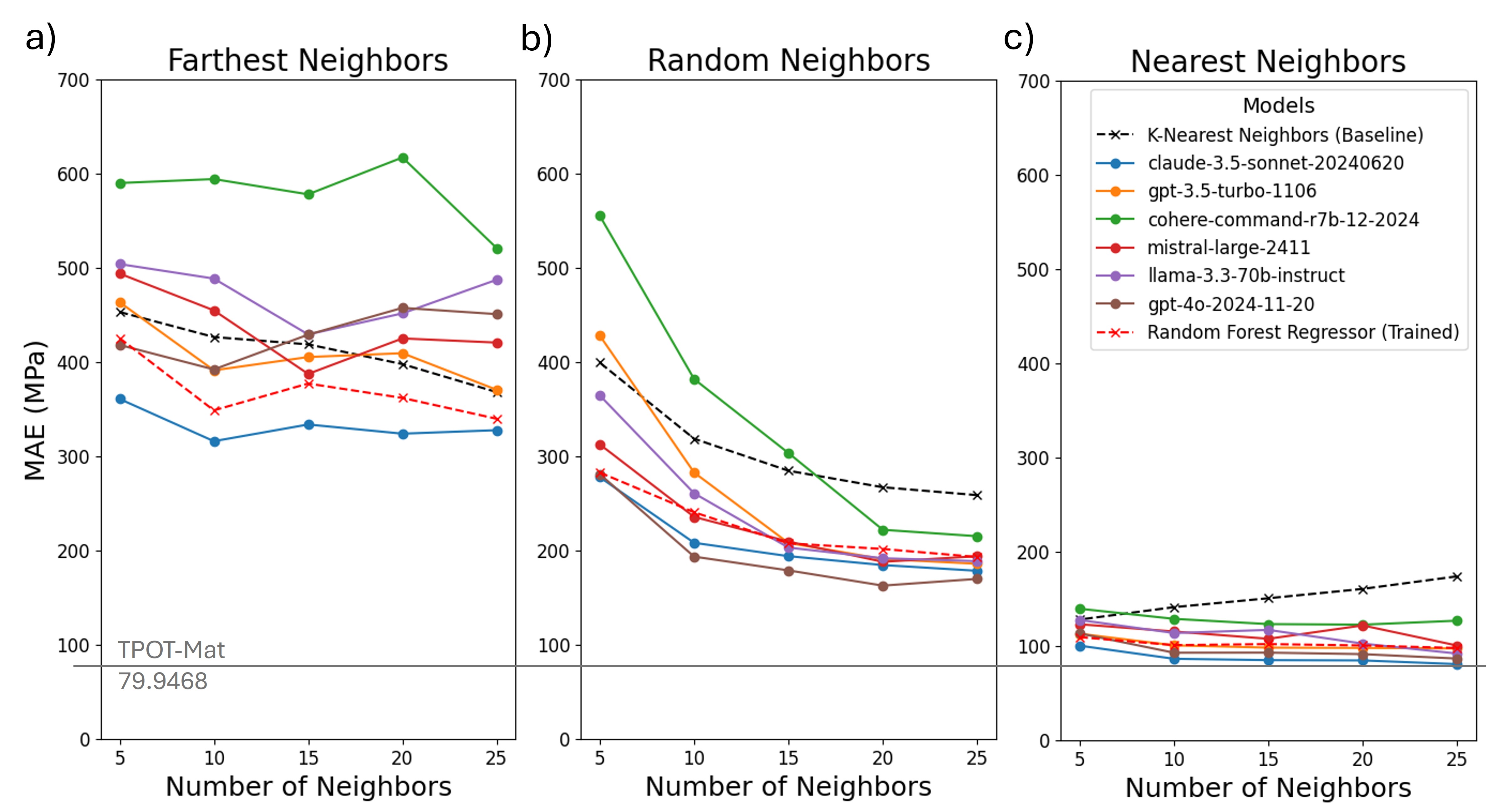}
\caption{Prediction performance of various LLMs under different training neighbor settings. Three panels compare the model performance across neighbor selection methods (left to right): a) Farthest Neighbors, b) Random Neighbors, and c) Nearest Neighbors. The TPOT-Mat performance is indicated by a horizontal grey line for benchmarking.}
\label{Revision-figures/Revision-llm_pred_all.jpg}
\end{figure*}

Here, we investigate LLM materials property prediction with ICL, utilizing the matbench\_steels dataset to predict the yield strength of steels. To explore how the selection of highly relevant training data can be used to enhance LLM performance in property prediction, we conducted a systematic study using nearest-neighbor-boosted ICL. The method developed here involves deliberately selecting training data points based on their representational proximity, meaning that data points with material compositions most similar to the test sample are used as few-shot examples to enhance the model's performance. Figure \ref{Revision-figures/Revision-llm_pred_all.jpg} shows the prediction performance when LLMs are tasked with predicting yield strength using a) farthest neighbors, b) random neighbors, and c) nearest neighbors. For performance comparison, a k-nearest neighbors (\textit{KNN}) model and a random forest regressor (\textit{RFR}) were also implemented. The \textit{RFR} is explicitly trained using MAGPIE features\cite{Ward2016} from the same data points used in ICL. While there is some correlation between material composition and yield strength, composition alone is not a strong predictor of yield strength. The KNN model serves as an additional baseline to demonstrate that the LLMs are not merely averaging or interpolating values from the selected examples but may be identifying implicit patterns in the data. The results are shown in Figure \ref{Revision-figures/Revision-llm_pred_all.jpg}. 

When trained with farthest neighbors, models generally exhibit high MAE with no clear trend as the number of neighbors increases. Most models perform worse than the \textit{KNN} and \textit{RFR} models except for \textit{claude-3.5-sonnet-20240620}, which slightly outperforms the \textit{RFR} but still exhibits high MAE. These results suggest that, when provided with distant data points, both LLMs and traditional ML models struggle to make valid predictions. This highlights a key challenge in OOD generalization, as training examples that are too dissimilar to the test sample prevent models from capturing meaningful structure-property relationships, leading to higher prediction errors.

For the random neighbors training set, the LLMs' performance consistently improves as the dataset size increases. This suggests that randomly composed few-shot examples offers a more balanced and diverse learning environment for these models, allowing models to develop more robust generalization. The \textit{claude-3.5-sonnet-20240620} and \textit{gpt-4o-2024-11-20} models consistently outperform the \textit{RFR} model as the data size increases, indicating that their more sophisticated architectures and larger training corpora enhance their ability to analyze and interpret complex data relationships more effectively. On the other hand, the smaller and older LLMs (i.e., \textit{cohere-command-r7b-12-2024} and \textit{gpt-3.5-turbo-1106}) exhibit higher MAE values throughout. The random neighbors setting appears to challenge these models to a greater degree, likely due to their smaller scale and pretraining, which limit their ability to generalize effectively to diverse inputs without fine-tuning or additional data processing. However, while their overall performance remains lower, their MAE decreases more significantly with more few-shot examples, suggesting that these LLMs can benefit from more context. 

The nearest neighbors represent the most relevant data points in the compositional space to the test points. As expected, the \textit{KNN} shows an increase in MAE as the number of neighbors grows, since additional (more distant) neighbors are less similar to the predicted sample. If LLMs relied solely on the provided information without additional internal computation, a similar performance decline would be expected. Contrastingly, as the data size increases, all the LLMs show a consistent decrease in MAE and outperform the \textit{KNN} model after 5 points. The more advanced LLMs (i.e., \textit{claude-3.5-sonnet-20240620} and \textit{gpt-4o-2024-11-20}) consistently outperform the trained \textit{RFR} model, suggesting that they can capture more complex relationships in the data rather than solely relying on interpolation from the provided examples. Notably, with 25 nearest neighbors, \textit{claude-3.5-sonnet-20240620} achieves an MAE of 80.5, nearly matching the best-performing ML model, TPOT-Mat, which achieves an MAE of 79.9 on the matbench\_steels dataset \cite{Olson2016}. However, it is important to note that this is not a direct comparison, as TPOT-Mat employs a 5-fold nested cross-validation method on Matbench datasets \cite{Dunn2020}, which utilizes 80\% of the data and is likely to result in better performance. The results highlight the potential of LLMs in data-lean materials property prediction tasks without the need for feature engineering, particularly given their general-purpose design and lack of task-specific fine-tuning. 

The results suggest that pretrained LLMs may exhibit adaptability to new predictive challenges using ICL, particularly when data availability is limited. While their ability to extract patterns from a small number of examples is promising, their performance remains task-dependent and may not generalize across all types of property predictions. A key insight is that LLMs can be potentially valuable in early-stage research or exploratory studies in materials science, where data may be scarce or costly to obtain. One potential use case is active learning, where LLMs help identify the most informative data points for experimental validation, optimizing the data acquisition process and reducing the number of required experiments while still achieving meaningful insights. However, as the number of data points increases, most LLMs suffer from limited prompt windows, which make such applications computationally expensive or impossible, in which case fine-tuned LLMs and traditional machine learning models with dedicated training may be more effective. 

\begin{figure*}
\centering
\includegraphics[height=9.2cm]{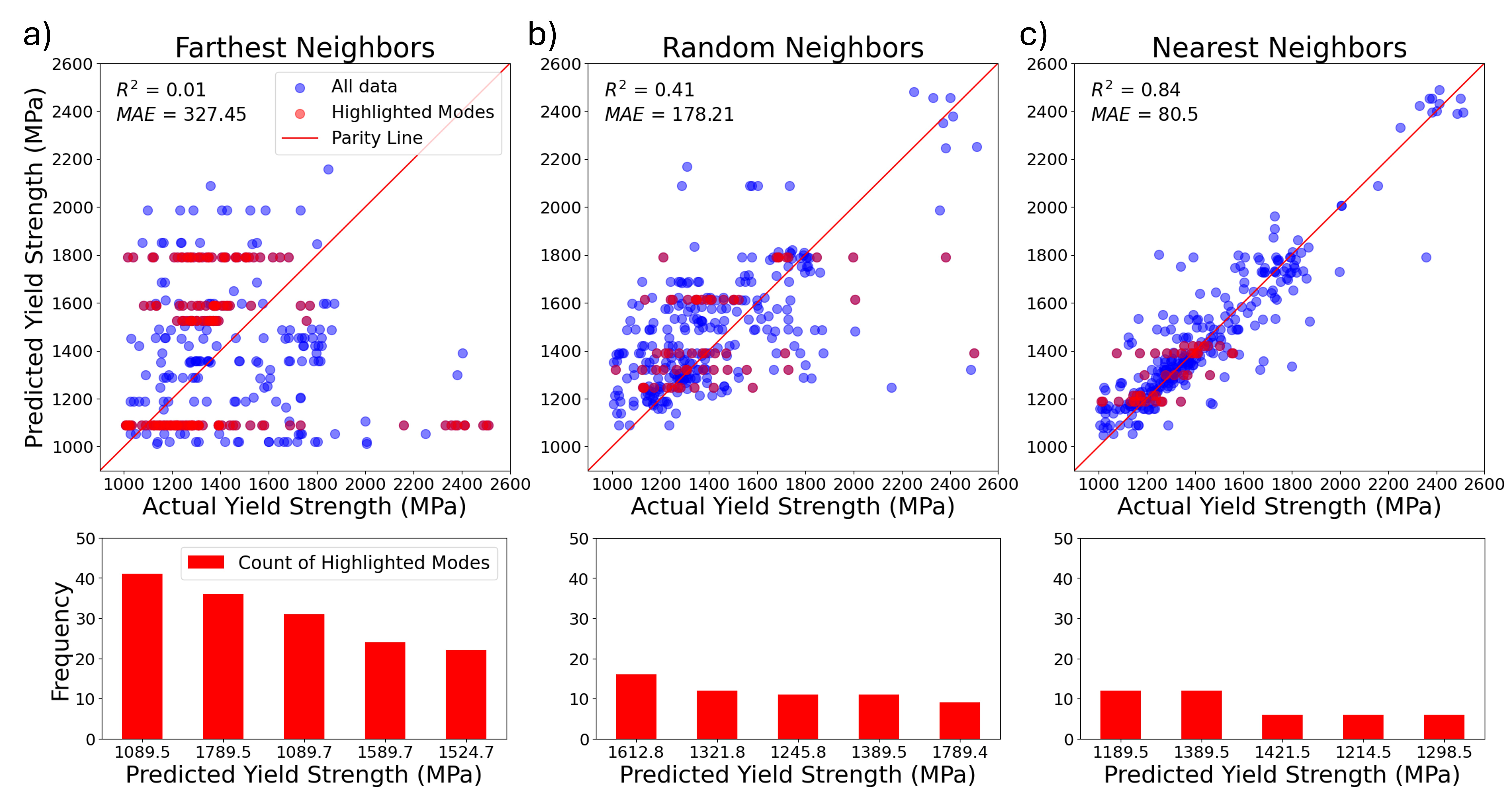}
\caption{Parity plots (top) and associated histograms (bottom) of highlighted modes of \textit{claude-3.5-sonnet-20240620} with 25 neighboring data points under different training neighbor settings: a) Farthest Neighbors, b) Random Neighbors, and c) Nearest Neighbors, from left to right.}
\label{Revision-figures/Revision-claude_pred.jpg}
\end{figure*}

To investigate the model’s predictive behaviors under these different settings, we analyzed the parity plots of the \textit{claude-3.5-sonnet-20240620}’s predictions when utilizing 25 neighboring data points, as shown in Figure \ref{Revision-figures/Revision-claude_pred.jpg}. Alongside these plots, the figure also includes histograms of the top five most frequently predicted yield strength values, to investigate whether the model is merely guessing a few commonly present values (shown as the red points in the parity plots). This behavior is known as “mode collapse”, whereby a generative model can favor a certain output due to overfitting to its pretraining data or lack of generalization capability \cite{thanhtung2020catastrophicforgettingmodecollapse}. Understanding mode collapse is crucial for evaluating the robustness of LLMs because it directly impacts the model's reliability and utility in practical applications. By identifying the mode collapse behavior, one can evaluate the validity of those predictions and potentially improve the performance. 

In the farthest neighbors setting, the red points in the figure form horizontal lines, indicating that the model frequently predicts the same yield strength values regardless of composition.This suggests that it fails to capture the underlying relationship between composition and yield strength effectively. The histogram further reveals a strong mode collapse behavior, with the model repeatedly predicting a set of values. This suggests that the model may be defaulting to a “safe” prediction range (possibly from the pretraining data) when provided with less relevant examples. This aligns with the shortcut learning behavior observed in LLMs, where models rely on superficial correlations rather than learning meaningful patterns from the data \cite{du2023shortcutlearninglargelanguage}. Instead of extrapolating from compositional trends, the model may be leveraging spurious cues from its training distribution, leading to repetitive and less informative predictions. In the random neighbors setting, the model shows better overall performance and a reduced mode collapse behavior. This suggests that introducing more variability into the training data helps the model to better understand the underlying patterns that predict yield strength. The nearest neighbors setting exhibits the best performance suggesting that higher training data proximity can lead to more accurate predictions. The mode collapse behavior is significantly reduced compared to the farthest neighbors and random neighbors, showing a greater diversity in the model's output. 

The observations show varying degrees of mode collapse based on the proximity of prompt examples to the test point. For instance, when provided with more closely related training data, the model exhibits stronger predictive signals, reinforcing its dependence on contextual cues rather than learned generalization. The results from Figure \ref{Revision-figures/Revision-llm_pred_all.jpg} and \ref{Revision-figures/Revision-claude_pred.jpg} suggest that LLMs do not develop an intrinsic understanding of structure-property relationships but instead rely heavily on contextual information from the prompt. While they can be effective in-distribution predictors when guided by well-curated few-shot examples, their utility in extrapolative property prediction remains limited. However, in the context of active learning, the mode collapse behavior could serve as an unsupervised indicator of prediction validity, allowing for a self-diagnostic approach to assess whether the model is making meaningful predictions. Since mode collapse manifests as a loss of predictive diversity, its prevalence across test samples may highlight regions of high model uncertainty where additional experimental validation is most needed.

\subsection{Robustness Analysis of Materials Science Q\&A}

\begin{figure}[h]
\centering
\includegraphics[width=\linewidth]{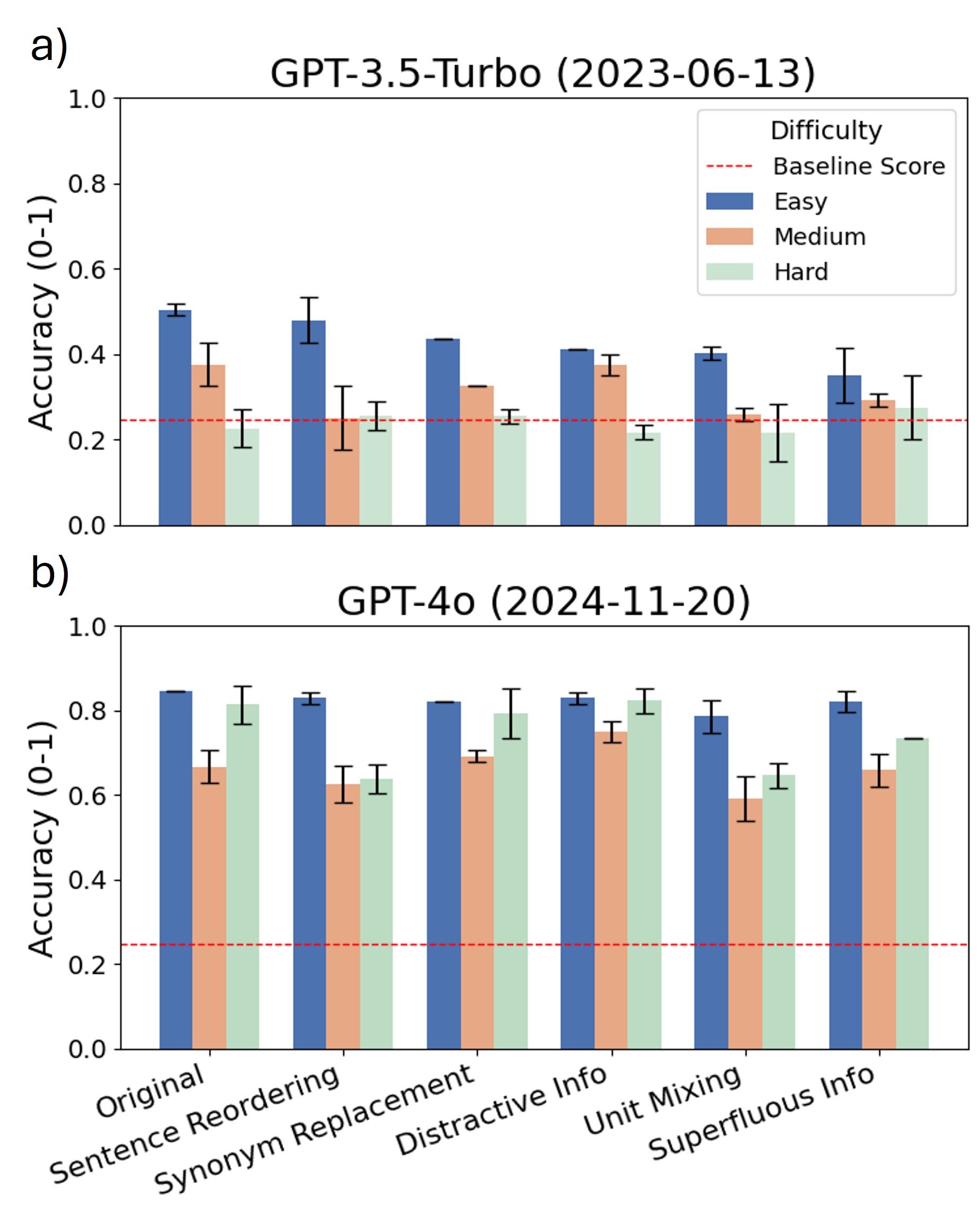}
\caption{Robustness analysis of a) \textit{gpt-3.5-turbo-0613} and b) \textit{gpt-4o-2024-11-20} in materials science Q\&A using MSE-MCQs dataset. The error bars represent the standard deviation due to LLMs’ non-determinism (Original, Synonym Replacement, Distractive Info) and the randomness (Sentence Reordering, Unit Mixing, Superfluous Info) introduced to the questions.}
\label{Revision-figures/Revision-llm_degrade_compareV.jpg}
\end{figure}

In Figure \ref{Revision-figures/Revision-llm_degrade_compareV.jpg}, we present the outcomes of the robustness assessment of \textit{gpt-3.5-turbo-0613} and \textit{gpt-4o-2024-11-20} when confronted with different types of textual modifications to the MSE-MCQs (see Table \ref{tab:degradation_types}), to evaluate its stability to various types of "noise". The comparison of these two models further demonstrates how the evolution of LLMs has influenced their robustness, contextual understanding, and overall reliability in processing modified inputs. 

Ranking by the degradation severity on the easy-level questions for \textit{gpt-3.5-turbo-0613}, sentence reordering has the least performance drop, followed by synonym replacement, distractive info, unit mixing, and superfluous info. The performance on the hard-level questions is close to the baseline score, indicating that \textit{gpt-3.5-turbo-0613} struggles with complex queries regardless of textual modifications, and will not be discussed in detail. The larger error bars in medium and hard questions suggest that LLMs tend to generate more varied responses to complex and lengthy queries. In contrast, the newer and more advanced model, \textit{gpt-4o-2024-11-20}, shows minimal degradation on the easy-level questions, maintaining an accuracy above 0.8, except for unit mixing. This suggests that the model is better at handling text changes and more robust than its predecessor. However, performance degradation becomes more noticeable on medium and hard questions. Sentence reordering, unit mixing, and superfluous info have the most significant impact on \textit{gpt-4o-2024-11-20} at these difficulty levels. 

Sentence reordering has little effect on the performance of both \textit{gpt-3.5-turbo-0613} and \textit{gpt-4o-2024-11-20} on easy-level questions, indicating that both models can effectively parse and extract key information even when the natural flow of a question is altered. However, the impact becomes more significant on medium and hard-level questions, where reordering appears to disrupt comprehension more significantly. This suggests that while both models exhibit strong syntactic flexibility in simpler cases, they may rely more heavily on common question structures when dealing with more complex queries.

The slight performance drop with synonym replacement suggests that both \textit{gpt-3.5-turbo-0613} and \textit{gpt-4o-2024-11-20} are somewhat sensitive to changes in terminology, leading to occasional inconsistencies in their responses. This reveals LLMs’ reliance on specific terminologies for recognition and comprehension in materials science. In contrast to humans, who can grasp the conceptual continuity behind varied expressions for flexible cognition, LLMs' struggles with synonym replacement emphasize the need for advanced training that prioritizes semantic networks over mere word recognition \cite{ye2024llmdadataaugmentationlarge}.

Introducing distractive information simulates a real-world scenario where irrelevant data often accompanies critical information, requiring sharp focus and analytical precision. Improving LLMs' ability to filter out irrelevant information is crucial for more effective information retrieval, problem-solving, and data interpretation \cite{wu2024easilyirrelevantinputsskew}. The results indicate that both models experience slight degradation on easy-level questions, where mostly simple conceptual knowledge is tested. The added non-essential details seemed to divert the LLMs' "attention", leading to fewer correct answers. Interestingly, distractive info seems to have improved performance on \textit{gpt-4o-2024-11-20} on medium and hard questions. This may be rationalized by the possibility that the added information inadvertently help the model by reinforcing key concepts or encouraging deeper contextual reasoning, functioning similarly to prompt engineering in guiding the model toward more accurate responses.

Mixing and converting the units tests LLMs’ abilities to perform numerical reasoning and apply mathematical concepts within a linguistic context. The added complexity introduced by unit mixing degraded the performance of both models, indicating challenges in handling numerical transformations embedded in text. Although some state-of-the-art LLMs support multi-modal applications and function calls to perform calculations \cite{wang2024mllmtoolmultimodallargelanguage}, accurately identifying and converting units within large text can still be critical. Improving this ability could enhance LLMs’ effectiveness in tasks such as information retrieval, data interpretation, and scientific analysis, where precise numerical reasoning is essential.

Superfluous information differs from distractive information in that it is more relevant to the questions themselves. The extent of performance degradation is likely influenced by the type and relevance of the superfluous information provided. The results show that \textit{gpt-3.5-turbo-0613} struggles significantly with superfluous information, experiencing the most severe performance degradation among all modifications. This suggests that it has difficulty filtering out non-essential details, leading to confusion or misinterpretation. In contrast, \textit{gpt-4o-2024-11-20} remains largely unaffected on easy and medium-level, but experiences moderate degradation on hard questions. This indicates that while \textit{gpt-4o-2024-11-20} demonstrates stronger information selection capabilities, its ability to filter out unnecessary details weakens as question complexity increases. For LLMs, distinguishing the necessary information from merely relevant but non-essential details is a more challenging cognitive process, mirroring advanced human problem-solving. It requires an understanding of the problem's objective, prioritizing information based on the question, and applying only the information that will lead to the correct conclusion. This highlights a potential area for improvement in LLMs, particularly in their ability to assess and prioritize critical information in complex reasoning tasks.

\subsection{Robustness Analysis of Materials Property Prediction}

Table \ref{tab:llm_prop_degrade} shows the result of the degradation study on the \textit{LLM-Prop} model, demonstrating how the LLM's performance on the band gap prediction is affected by various modifications to the textual descriptions of material crystal structures.

\begin{table*}[t]  
\footnotesize
\centering
\renewcommand{\arraystretch}{1.2} 
\caption{Mean Absolute Error (MAE) under different conditions}
\label{tab:llm_prop_degrade}
\resizebox{\textwidth}{!}{  
\begin{tabular}{lcccc} 
    \hline
    & \textbf{Original} & \textbf{Distractive Info} & \textbf{Sentence Reordering} & \textbf{Misleading Info} \\
    \hline
    \textbf{MAE (eV)} & 0.286 & 0.287 & $0.323 \pm 0.002$ & $0.398 \pm 0.005$ \\
    \hline
\end{tabular}
}
\end{table*}

After adding distractive information to the material descriptions, the \textit{LLM-Prop} model showed negligible degradation, indicating this application-specific model can effectively differentiate relevant from irrelevant information. This resilience, likely due to the targeted training and fine-tuning on domain-specific texts, enables it to focus on key features for band gap prediction. This showcases the potential noise-filtering capabilities of the trained and fine-tuned transformer models, which traditional ML models may suffer from. 

The impact of sentence reordering increased the MAE by 12.9\%, suggesting the model's reliance on the structured descriptions for accurate predictions. From the previous study on MSE-MCQs degradation, the effect of sentence reordering was less significant, indicating that larger general LLMs, which are trained with more various texts, can exhibit better contextual understanding and are less prone to order changes.

The presence of misleading information, particularly an additional sentence from another material's description, leads to a 39\% increase in MAE. This substantial degradation indicates that while the model can filter out irrelevant distractive noise, it struggles considerably when faced with data that is contextually relevant to the specific prediction task. Notably, this impact arises from the addition of just a single misleading sentence, highlighting the model’s vulnerability to subtle contextual inconsistencies that misdirect its predictions.

\begin{figure}[h]
\centering
    \includegraphics[width=\linewidth]{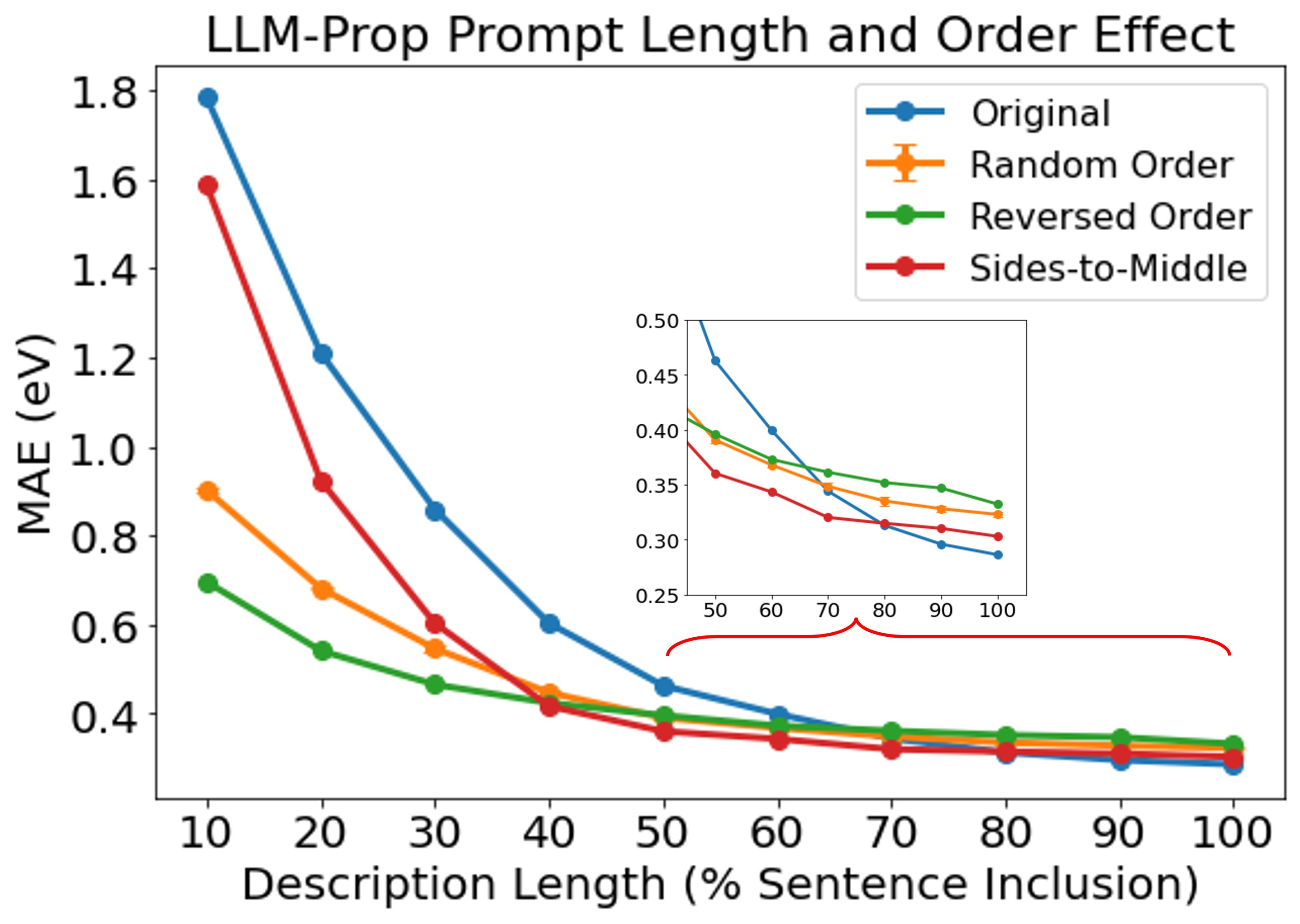}
    \caption{Oreder and length effect of \textit{LLM-Prop} on prediction performance.}
    \label{Revision-figures/llm-prop.jpg}
\end{figure}

To further assess the model's robustness and determine which description elements are essential for prediction accuracy, we conducted a truncation study that involves altering the orders and lengths of the input description. As shown in Figure \ref{Revision-figures/llm-prop.jpg}, the description length is expressed as percent sentence inclusion, ranging from 10\% to 100\% and MAE is used as a measure of prediction accuracy. 

When the number of description sentences is incrementally increased, the MAE rapidly decreases and minimized at 100\% sentence inclusion. Interestingly, in the random order, reversed order, and sides-to-middle configurations, the initial MAE at 10\% sentence inclusion is notably lower than in the original order, with some configurations achieving nearly double the performance of the original setting. This indicates that the initial sentences may not contain the most useful information for prediction. The MAEs begin to converge around 50\% sentence inclusion, beyond which differences become statistically insignificant. Notably, in the random order setting, there is virtually no variation in the MAE when incorporating three different sentence shuffles. This suggests that \textit{LLM-Prop} can effectively extract key information and deliver consistent predictions, despite variations in the sentence order. 

By 40\% sentence inclusion, the reversed order yields the lowest MAE, indicating that sentences at the end of descriptions contain crucial predictive information. However, by 50\% sentence inclusion, the performance of the reversed order begins to align with that of the random order, suggesting that central information in the descriptions may not be as crucial for prediction accuracy. Since random order includes more initial sentences than reversed order, this suggests that the first sentences may contribute less relevant details, particularly at lower inclusion percentages. Based on these insights, we developed the sides-to-middle approach, aiming to prioritize information at the beginning and the end. This approach consistently outperforms other configurations between 40\% and 70\% sentence inclusion, achieving the lowest MAE in this range. The error continues to decrease and is optimized at full sentence inclusion being only 5.8\% higher than the original setting in MAE. This result suggests that while the original order remains optimal, the contextual framing provided by the beginning and end of descriptions is particularly important for model accuracy.

This truncation study showcases that the fine-tuned model can perform effectively even when provided with significantly reduced prompts. We found that diverging from the training setup (i.e., changing the textual order of the prompt) can sometimes result in improved performance at truncated data volumes. This counterintuitive result suggests that highly templated training or fine-tuning data can lead to unexpected effects. Consequently, this implies two key considerations: 1) training templates should be diverse to prevent models from overfitting to unimportant patterns, and 2) when using a fine-tuned model trained on a specific template, it may not always be optimal to match the template during inference. These insights highlight the potential for optimizing training costs while maintaining performance.

\section*{Conclusions}
The findings of this study offer crucial insights into the behaviors and limitations of LLMs in materials science domain-specific Q\&A and materials property prediction. While the robustness analysis indicates that LLMs can manage certain types of noise with resilience, their performance is significantly challenged under more complex and deceptive conditions, such as when superfluous information is introduced. In Q\&A tasks, prompting techniques (e.g., expert prompting, zero-shot chain-of-thought prompting) can sometimes improve the model performance in handling more complex queries, not by unlocking new capabilities, but by increasing the probability of following an expected format. In materials property predictions, we find that in-context learning allows pretrained LLMs to achieve relatively high accuracy in low-data settings when provided with few-shot examples with high proximity to the target material. However, the observed mode collapse behavior, where the model generates repeated outputs despite varying inputs, showcases that providing ineffective few-shot examples (i.e., out-of-distribution data points to the target material) can cause the model to default to a memorized response rather than conditioning its output on the provided prompt. This study also highlights that fine-tuned models can exhibit enhanced performance under truncated data conditions when diverging from the training setup, such as altering the textual order. While this suggests an unexpected level of robustness, it also exposes the risks associated with fixed templating during fine-tuning, suggesting that users should be cautious about strictly matching the training templates during inference. This study highlights the challenges and limitations of using LLMs in materials science, emphasizing the importance of better dataset curation, dynamic prompting techniques, and training strategies to enhance LLMs as reliable tools for materials discovery and scientific research.

\section*{Author contributions}
H.W., E.K., and J.H.S. conceived and designed the project. E.K., and J.H.S. supervised the project. H.W. conducted the experiments. S.R. provided the dataset for materials science question \& answering study. H.W., K.L., Y.F., E.K., and J.H.S. discussed the results. H.W drafted the manuscript. All authors reviewed and edited the manuscript.

\section*{Conflicts of interest}
The authors declare no conflicts of interest.

\section*{Data and code availability}
The datasets and code for the analyses and figure generations in this work are publicly available on GitHub at url: https://github.com/Toniaac/LLM-MSE-Eval-Robustness.git (DOI: 10.5281/zenodo.15007556). Detailed information on data and code access is provided in the README.md file in the GitHub repository.

\section*{Acknowledgments}
We acknowledge Andre Niyongabo Rubungo and Adji Bousso Dieng for providing the dataset and model for the \textit{LLM-Prop} study.

\begingroup
\footnotesize
\bibliography{main} 
\bibliographystyle{rsc} 
\endgroup
\end{document}


\section*{Supplemental Information}

\subsection*{Examples of MSE-MCQs questions and difficulty labeling}
\begin{table}[H] 
\centering
\resizebox{\textwidth}{!}{ 
\scriptsize 
\begin{tabularx}{\textwidth}{l|X}  
\hline
\textbf{Difficulty Label} & \textbf{MSE-MCQs Question} \\ 
\hline
Easy & 
Which of the following most closely describes the ductility of a sample?
\newline(a) The plastic strain at fracture
\newline(b) The elastic strain at fracture
\newline(c) The total strain at fracture
\newline(d) None of the above \\
\hline
Medium &
An hypothetical FCC metal has a density of 7.4 g/cm3 and a molar mass of 55.3 g/mol. Which of the following is the correct number of atom sites (that is, without any vacancies)?
\newline(a) 1.09e+22 atoms/cm3
\newline(b) 1.34e-01 atoms/cm3
\newline(c) 6.80e-22 atoms/cm3
\newline(d) 8.06e+22 atoms/cm3 \\
\hline
Hard &
A cylindrical sample of stainless steel having a Young's modulus of 204.3 GPa, a diameter of 12.0 mm, and initial length of 237.8 mm is loaded to a stress of 411.5 MPa. The sample is then completely unloaded. What will the elastic recovery of this sample be, in mm? The yield strength and ultimate tensile strength of this specific alloy are 292.0 MPa and 688.0 MPa, respectively.
\newline(a) Possible to calculate from information provided, but none of these options are correct.
\newline(b) 0.96
\newline(c) Not possible to calculate from information provided.
\newline(d) 239.0
\newline(e) 0.24
\newline(f) 0.48 \\ 
\hline
\end{tabularx}
}
\end{table}

\subsection*{Examples of textual perturbations of MSE-MCQs questions and crystal structure descriptions from LLM-Prop band gap dataset}
\begin{table}[H] 
\centering
\resizebox{\textwidth}{!}{ 
\scriptsize 
\begin{tabularx}{\textwidth}{l|X|X}  
\hline
\textbf{Degradation Type} & \textbf{MSE-MCQs Example Prompt} & \textbf{LLM-Prop Example Prompt} \\ 
\hline
Original & 
32 g of solid sugar is added to 68 g of liquid water at room temperature and allowed to reach equilibrium. No solid sugar is observed within the container. The temperature is then lowered to a temperature T1 and allowed to reach equilibrium. At temperature T1, it is observed that only solid sugar and solid water (ice) exist. There is no solubility of sugar in solid water, nor of water in solid sugar. How many grams of ice are present in this container?  
\newline(a) 68 g  
\newline(b) 32 g  
\newline(c) 36 g  
\newline(d) None of these options  
& Be\textsubscript{4}AlMn crystallizes in the cubic $\overline{F}43m$ space group. Be is bonded to six equivalent Be, three equivalent Mn, and three equivalent Al atoms to form a mixture of edge, face, and corner-sharing BeMn\textsubscript{3}Be\textsubscript{6}Al\textsubscript{3} cuboctahedra. There are three shorter (2.12 Å) and three longer (2.17 Å) Be–Be bond lengths. All Be–Mn bond lengths are 2.51 Å. All Be–Al bond lengths are 2.52 Å. Mn is bonded in a 16-coordinate geometry to twelve equivalent Be and four equivalent Al atoms. All Mn–Al bond lengths are 2.63 Å. Al is bonded in a 16-coordinate geometry to twelve equivalent Be and four equivalent Mn atoms. \\ 
\hline
Unit Mixing & 
\textcolor{red}{0.032 kg} of solid sugar is added to \textcolor{red}{0.068 kg} of liquid water at room temperature and allowed to reach equilibrium. No solid sugar is observed within the container. The temperature is then lowered to a temperature T1 and allowed to reach equilibrium. At temperature T1, it is observed that only solid sugar and solid water (ice) exist. There is no solubility of sugar in solid water, nor of water in solid sugar. How many grams of ice are present in this container?  
\newline(a) \textcolor{red}{0.150 lbs}  
\newline(b) \textcolor{red}{32000 mg} 
\newline(c) \textcolor{red}{36000 mg}  
\newline(d) None of these options  
& N/A \\ 
\hline
Sentence Reordering &  
\textcolor{red}{At temperature T1, it is observed that only solid sugar and solid water (ice) exist.} No solid sugar is observed within the container. The temperature is then lowered to a temperature T1 and allowed to reach equilibrium. \textcolor{red}{There is no solubility of sugar in solid water, nor of water in solid sugar.} \textcolor{red}{32 g of solid sugar is added to 68 g of liquid water at room temperature and allowed to reach equilibrium.} How many grams of ice are present in this container?  
\newline(a) 68 g  
\newline(b) 32 g  
\newline(c) 36 g  
\newline(d) None of these options  
& Be\textsubscript{4}AlMn crystallizes in the cubic $\overline{F}43m$  space group. \textcolor{red}{All Be–Al bond lengths are 2.52 Å}. \textcolor{red}{All Be–Mn bond lengths are 2.51 Å}. \textcolor{red}{All Mn–Al bond lengths are 2.63 Å.} \textcolor{red}{Mn is bonded in a 16-coordinate geometry to twelve equivalent Be and four equivalent Al atoms.} \textcolor{red}{There are three shorter (2.12 Å) and three longer (2.17 Å) Be–Be bond lengths.} \textcolor{red}{Al is bonded in a 16-coordinate geometry to twelve equivalent Be and four equivalent Mn atoms.} \textcolor{red}{Be is bonded to six equivalent Be, three equivalent Mn, and three equivalent Al atoms to form a mixture of edge, face, and corner-sharing BeMn\textsubscript{3}Be\textsubscript{6}Al\textsubscript{3} cuboctahedra.} \\ 
\hline
Synonym Replacement &  
32 g of solid \textcolor{red}{sucrose} is added to 68 g of \textcolor{red}{H$_2$O} at \textcolor{red}{ambient conditions} and allowed to reach equilibrium. No solid \textcolor{red}{sucrose} is observed within the container. The temperature is then reduced to a temperature T1 and allowed to reach equilibrium. At temperature T1, it is observed that only solid \textcolor{red}{sucrose} and \textcolor{red}{crystalline water} (ice) exist. There is no \textcolor{red}{dissolution} of \textcolor{red}{sucrose} in \textcolor{red}{crystalline water}, nor of water in solid \textcolor{red}{sucrose}. How many grams of ice are present in this container?  
\newline(a) 68 g  
\newline(b) 32 g  
\newline(c) 36 g  
\newline(d) None of these options  
& N/A \\ 
\hline
Distractive Information &  
\textcolor{red}{One morning, the mothers and fathers were ready as usual in the main square of Four Houses waiting for the red bus. But this was no ordinary day, for the red bus was twenty minutes late. Very slowly, he came chugging along but once he got to the square he collapsed from exhaustion. There was no way he could take the children to school today. Then the teacher asked:}  

32 g of solid sugar is added to 68 g of liquid water at room temperature and allowed to reach equilibrium. No solid sugar is observed within the container. The temperature is then lowered to a temperature T1 and allowed to reach equilibrium. At temperature T1, it is observed that only solid sugar and solid water (ice) exist. There is no solubility of sugar in solid water, nor of water in solid sugar. How many grams of ice are present in this container?  
\newline(a) 68 g  
\newline(b) 32 g  
\newline(c) 36 g  
\newline(d) None of these options  
& \textcolor{red}{One morning, the mothers and fathers were ready as usual in the main square of Four Houses waiting for the red bus. But this was no ordinary day, for the red bus was twenty minutes late. Very slowly, he came chugging along but once he got to the square he collapsed from exhaustion. There was no way he could take the children to school today. Then the teacher asked: }Be\textsubscript{4}AlMn crystallizes in the cubic $\overline{F}43m$  space group. Be is bonded to six equivalent Be, three equivalent Mn, and three equivalent Al atoms to form a mixture of edge, face, and corner-sharing BeMn\textsubscript{3}Be\textsubscript{6}Al\textsubscript{3} cuboctahedra. There are three shorter (2.12 Å) and three longer (2.17 Å) Be–Be bond lengths. All Be–Mn bond lengths are 2.51 Å. All Be–Al bond lengths are 2.52 Å. Mn is bonded in a 16-coordinate geometry to twelve equivalent Be and four equivalent Al atoms. All Mn–Al bond lengths are 2.63 Å. Al is bonded in a 16-coordinate geometry to twelve equivalent Be and four equivalent Mn atoms. \\ 
\hline
Superfluous Information&  
32 g of solid sugar is added to 68 g of liquid water at room temperature and allowed to reach equilibrium. \textcolor{red}{Under this load, the gauge length elongates elastically by 46 mm.} No solid sugar is observed within the container. The temperature is then lowered to a temperature T1 and allowed to reach equilibrium. At temperature T1, it is observed that only solid sugar and solid water (ice) exist. There is no solubility of sugar in solid water, nor of water in solid sugar. How many grams of ice are present in this container?  
\newline(a) 68 g  
\newline(b) 32 g  
\newline(c) 36 g  
\newline(d) None of these options  
& Be\textsubscript{4}AlMn crystallizes in the cubic $\overline{F}43m$  space group. \textcolor{red}{The corner-sharing octahedral tilt angles range from 19–44°.} Be is bonded to six equivalent Be, three equivalent Mn, and three equivalent Al atoms to form a mixture of edge, face, and corner-sharing BeMn\textsubscript{3}Be\textsubscript{6}Al\textsubscript{3} cuboctahedra. There are three shorter (2.12 Å) and three longer (2.17 Å) Be–Be bond lengths. All Be–Mn bond lengths are 2.51 Å. All Be–Al bond lengths are 2.52 Å. Mn is bonded in a 16-coordinate geometry to twelve equivalent Be and four equivalent Al atoms. All Mn–Al bond lengths are 2.63 Å. Al is bonded in a 16-coordinate geometry to twelve equivalent Be and four equivalent Mn atoms.  \newline(\textbf{Note: This information is misleading rather than superfluous}) \\
\hline
\end{tabularx}
}
\end{table}

\subsection*{Matbench\_steels dataset PCA and correlation analysis}
PC1\_Correlation = -0.145
\\PC2\_Correlation = 0.436
\begin{figure}[ht]
\centering
\includegraphics[width=\linewidth]{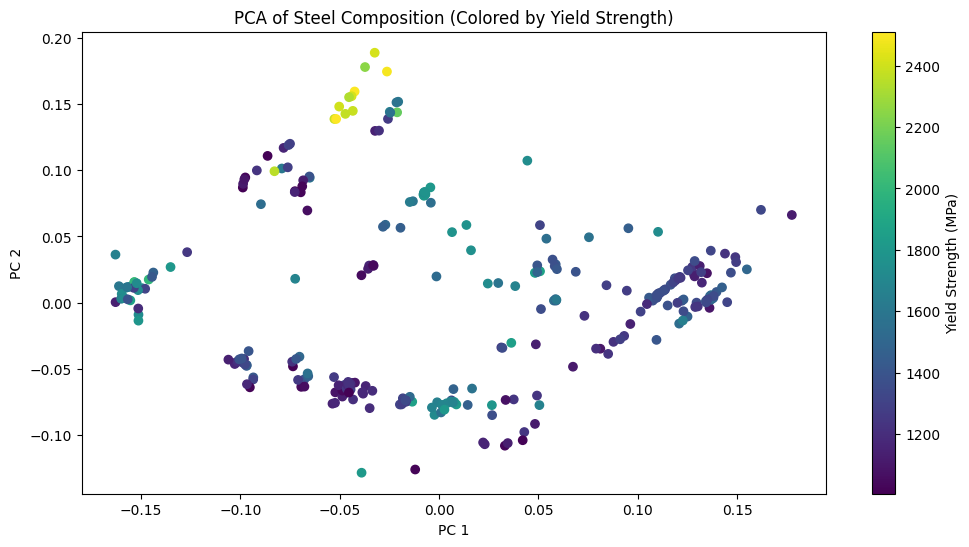}
\end{figure}

\subsection*{MAGPIE Features used to train the RFR model}
The following MAGPIE features were used to train the \textit{RFR} model:
\begin{itemize}
    \item Features: 
    \begin{itemize}
        \item \texttt{Number}, \texttt{MendeleevNumber}, \texttt{AtomicWeight}, \texttt{MeltingT}, \texttt{Column}, \texttt{Row}
        \item \texttt{CovalentRadius}, \texttt{Electronegativity}, 
        \item \texttt{NsValence}, \texttt{NpValence}, \texttt{NdValence}, \texttt{NfValence}, \texttt{NValence}
        \item \texttt{NsUnfilled}, \texttt{NpUnfilled}, \texttt{NdUnfilled}, \texttt{NfUnfilled}, \texttt{NUnfilled}
        \item \texttt{GSvolume\_pa}, \texttt{GSbandgap}, \texttt{GSmagmom}
    \end{itemize}
    \item Statistical Descriptors:
    \begin{itemize}
        \item \texttt{mean}, \texttt{avg\_dev}, \texttt{minimum}, \texttt{maximum}, \texttt{range}
    \end{itemize}
\end{itemize}

\subsection*{Example prompt of crystal structure descriptions from LLM-Prop band gap dataset}
\begin{figure}[H]
\centering
\includegraphics[width=\linewidth]{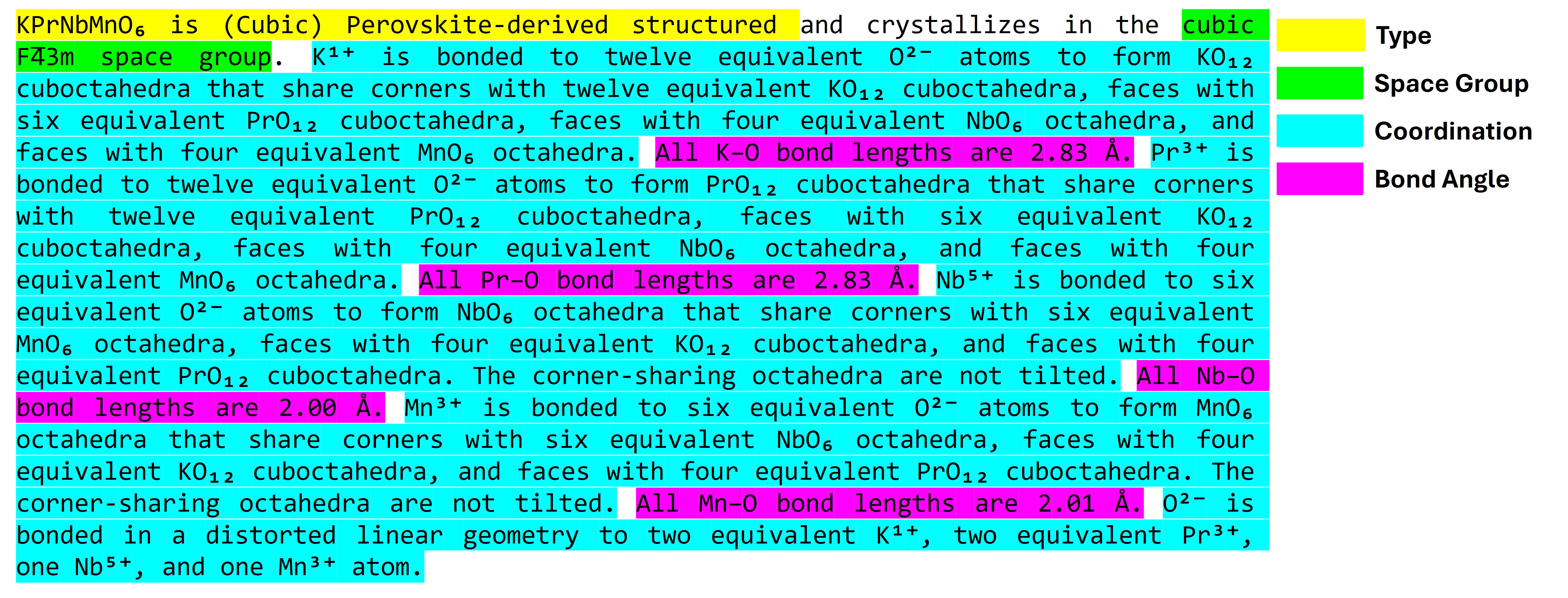}
\end{figure}